\documentclass[runningheads]{llncs}
\usepackage{graphicx}

\usepackage[dvipsnames]{xcolor}
\usepackage{tikz}
\usepackage{comment}
\usepackage{amsmath,amssymb} 

\usepackage[utf8]{inputenc}
\usepackage{booktabs}       
\usepackage{amsfonts}       
\usepackage{url}            
\usepackage{nicefrac}       
\usepackage{microtype}      
\usepackage{multirow}
\usepackage{multicol}
\usepackage{xspace}
\usepackage{adjustbox}
\usepackage{import}
\usepackage{subcaption}
\usepackage{color}
\usepackage{colortbl}
\usepackage{framed}
\usepackage{makecell}
\usepackage{wrapfig}
\usepackage{enumitem}
\usepackage{wasysym}

\usepackage{bbding}
\usepackage{pifont}
\usepackage{listings}
\usepackage{tabulary}
\usepackage{diagbox}
\usepackage{graphicx}
\usepackage[normalem]{ulem}

\usepackage[accsupp]{axessibility}  

\definecolor{OliveGreen}{rgb}{0.0, 0.55, 0.0}
\usepackage[pagebackref=false,breaklinks=true,letterpaper=true,colorlinks, citecolor = OliveGreen, bookmarks=false]{hyperref}

\newcommand*\samethanks[1][\value{footnote}]{\footnotemark[#1]}

\definecolor{shadecolor}{rgb}{0.92,0.92,0.92}

\newcommand{\cmark}{\ding{51}}%
\newcommand{\xmark}{\ding{55}}%

\begin{document}
\pagestyle{headings}
\mainmatter
\def\ECCVSubNumber{1398}  

\title{Expanding Language-Image Pretrained Models for General Video Recognition}


\titlerunning{X-CLIP}
%
\author{
    Bolin Ni\textsuperscript{1,2 \thanks{Work done during internship at Microsoft Research. $\dagger$ Project Lead.}},
    Houwen Peng\textsuperscript{1 $\dagger$},
    Minghao Chen\textsuperscript{1,3 \samethanks},
    Songyang Zhang\textsuperscript{4}, \\
    Gaofeng Meng\textsuperscript{2}, 
    Jianlong Fu\textsuperscript{1},
    Shiming Xiang\textsuperscript{2},
    Haibin Ling\textsuperscript{3}
}
\authorrunning{B. Ni et al.}
%
\institute{\quad\quad ~~$^{1}$ Microsoft Research \quad \quad \quad  \; $^{2} $ Chinese Academy of Sciences\quad \quad \quad  \\
$^{3}$ Stony Brook University \quad \quad \quad  \; $^{4} $ University of Rochester\\}

\maketitle

\begin{abstract}
Contrastive language-image pretraining has shown great success in learning visual-textual joint representation from web-scale data, demonstrating remarkable “zero-shot” generalization ability for various image tasks. However, how to effectively expand such new language-image pretraining methods to video domains is still an open problem. In this work, we present a simple yet effective approach that adapts the pretrained language-image models to video recognition directly, instead of pretraining a new model from scratch. 
More concretely, to capture the long-range dependencies of frames along the temporal dimension, we propose a cross-frame attention mechanism that explicitly exchanges information across  frames. Such module is lightweight and can be plugged into pretrained language-image models  seamlessly. 
\textcolor{black}{Moreover, we propose a video-specific prompting scheme, which leverages video content information for generating discriminative textual prompts.} Extensive experiments demonstrate that our approach is effective and can be  generalized to different video recognition scenarios. In particular, under fully-supervised settings, our approach achieves a top-1 accuracy of \textcolor{black}{87.1\%} on Kinectics-400, while using 12× fewer FLOPs compared with Swin-L and ViViT-H. In zero-shot experiments, our approach surpasses the current state-of-the-art methods by \textcolor{black}{+7.6\%} and \textcolor{black}{+14.9\%} in terms of top-1 accuracy under two popular protocols. In few-shot scenarios, our approach outperforms previous best methods by \textcolor{black}{+32.1\%} and \textcolor{black}{+23.1\%} when the labeled data is extremely limited. Code and models are available at \href{https://github.com/microsoft/VideoX/tree/master/X-CLIP}{aka.ms/X-CLIP}.
\keywords{Video Recognition, Contrastive Language-Image Pretraining}
\end{abstract}

\section{Introduction}

\begin{figure}
  \centering
  \includegraphics[scale=0.25]{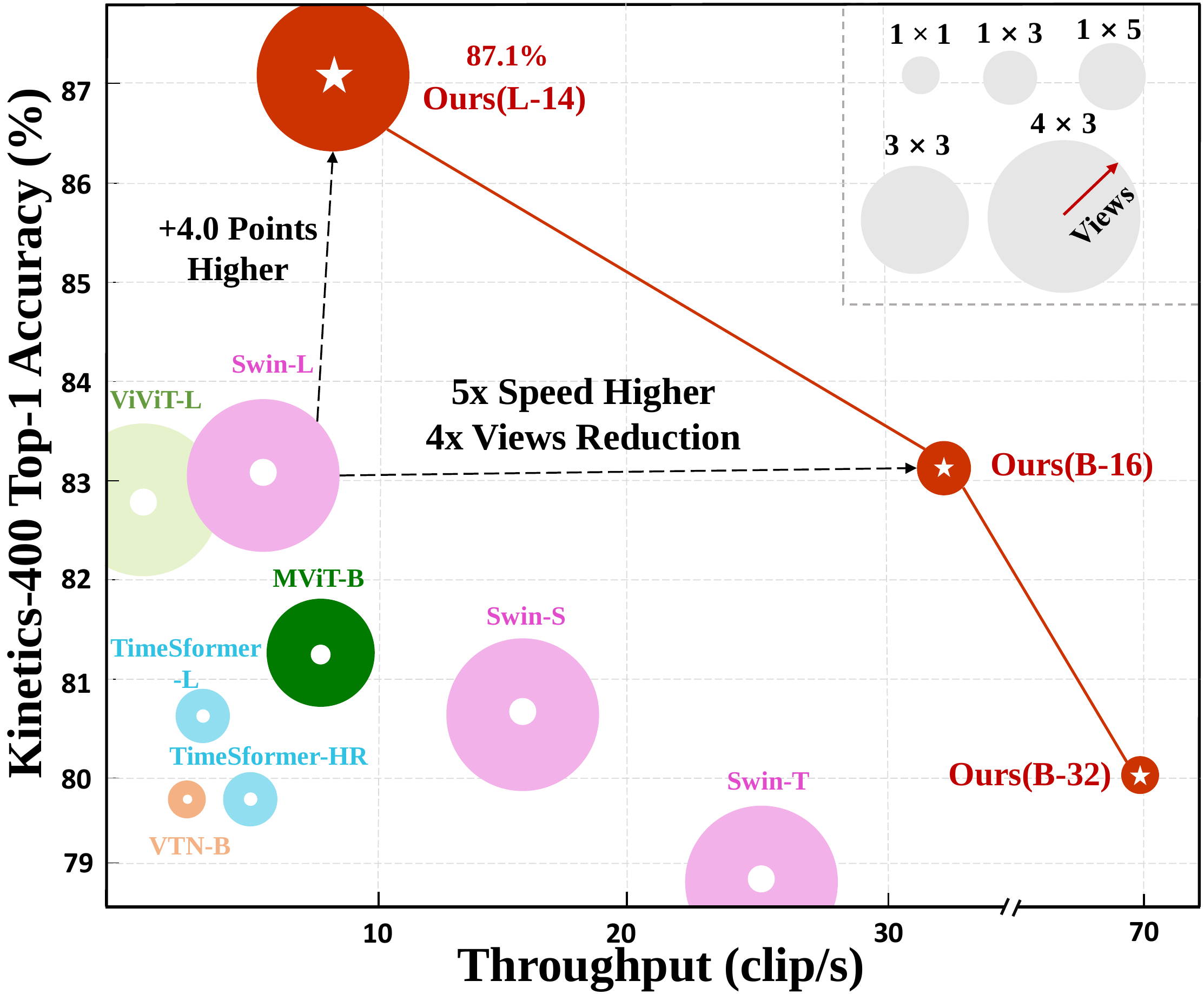}
  \includegraphics[scale=0.359]{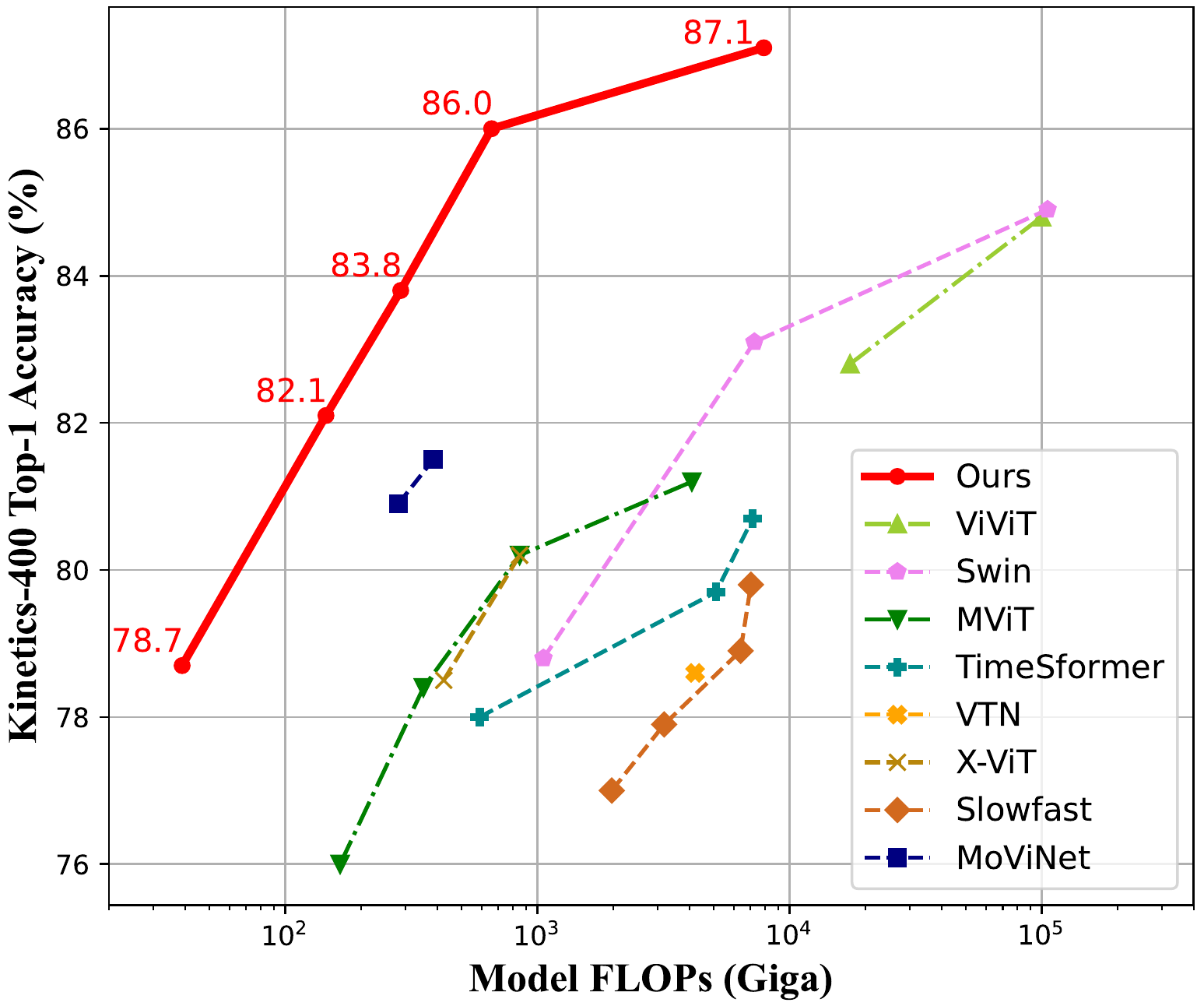}
  \caption{Comparison with state-of-the-art methods on Kinetics-400 \cite{k400} in terms of throughput, the number of views, and FLOPs. Best viewed in color.
  }
  \label{fig:throughputs_flops}
\end{figure}

Video recognition is one of the most fundamental yet challenging tasks in video understanding. It plays a vital role in numerous vision applications, such as micro-video recommendation~\cite{survey_action}, sports video analysis~\cite{survey_videotransformer}, autonomous driving~\cite{survey_action2}, and so on. Over the past few years, based upon convolutional neural networks and now transformers, video recognition has achieved remarkable progress \cite{survey_action,karpathy2014large}. Most existing works follow a \textcolor{black}{closed-set} learning setting, where all the categories are pre-defined. Such method is unrealistic for many real-world applications, such as automatic tagging of web videos, where information regarding new video categories is not available during training. It is thus very challenging for \textcolor{black}{closed-set} methods to train a classifier for recognizing unseen or unfamiliar categories.

Fortunately, recent work in large-scale contrastive language-image pretraining, such as CLIP \cite{clip}, ALIGN \cite{align}, and Florence \cite{florence}, has shown great potentials in addressing this challenge. The core idea is to learn visual or visual-language representation with natural language supervision using web-scale image-text data. \textcolor{black}{After pretraining, natural language is used to reference learned visual concepts (or describe new ones), thus enabling zero/few-shot transfer of the models to downstream tasks.} Inspired by these works \cite{clip,align,florence}, we consider to use text as the supervision signals to learn a new video representation for general recognition scenarios, including zero-shot, few-shot, and fully-supervised.

However, directly training a language-video model is unaffordable for many of us, because it requires large-scale video-text pretraining data as well as a massive number of GPU resources (\emph{e.g.}, thousands of GPU days). A feasible solution is to adapt the pretrained language-image models to video domain. Very recently, there are several studies exploring how to transfer the knowledge from the pretrained language-image models to other downstream tasks, \emph{e.g.}, point cloud understanding \cite{zhang2021pointclip} and dense prediction \cite{rao2021denseclip,zhou2021denseclip}. However, the transfer and adaptation to video recognition is not well explored. When adapting the pretrained cross-modality models from image to video domain, there are two key issues to be solved: 1) how to leverage the temporal information contained in videos, and 2) how to acquire discriminative text representation for a video.

For the first question, we present a new architecture for video temporal modeling. It consists of two key components: a cross-frame communication transformer and a multi-frame integration transformer. Specifically, the cross-frame communication transformer takes raw frames as input and provides a frame-level representation using a pretrained language-image model, while allowing information exchange between frames with message tokens. Each message token not only depicts the semantics of the current frame, but also communicates with other frames to model their dependencies. The multi-frame integration transformer then simply transfer the frame-level representations to video-level.

For the second question, we employ the text encoder pretrained in the language-image models and expand it with a video-specific prompting scheme. The key idea is to leverage video content information to enhance text prompting. The intuition behind is that appropriate contextual information can help the recognition. For example, if there is extra video content information about ``\texttt{in the water}", the actions ``\texttt{swimming}" and ``\texttt{running}" will be much easier to be distinguished. 
In contrast to prior work manually designing a fixed set of text prompts, this work proposes a learnable prompting mechanism, which integrates both semantic labels and representation of videos for automatic prompt generation.

With the above two issues addressed, we can smoothly adapt existing image-level cross-modality pretrained models to video domains. \textcolor{black}{Without loss of generality, here we choose the CLIP \cite{clip} and Florence \cite{florence} models and eXpand them for general video recognition, forming new model families called X-CLIP and X-Florence, respectively.} Comprehensive experiments demonstrate our expanded models are generally effective. In particular, under the fully-supervised setting, X-CLIP-L/14 achieves \textcolor{black}{competitive performance on Kinetics-400/600 with top-1 accuracies of 87.1\%/88.3\%, surpassing ViViT-H~\cite{arnab2021vivit} by $2.3\%/2.5\%$ while using 12× fewer FLOPs, as shown in Fig.~\ref{fig:throughputs_flops}}. In zero-shot experiments, \textcolor{black}{X-Florence} surpasses the state-of-the-art ActionCLIP \cite{wang2021actionclip} by +7.6\% and +14.9\% under two popular protocols. In few-shot experiments, X-CLIP outperforms other prevailing methods by $+32.1\%$ and $+23.1\%$ when the data is extremely limited.

In summary, our contributions are three-fold: 
\begin{itemize}
    \item We propose a new cross-frame communication attention for video temporal modeling. This module is light and efficient, and can be seamlessly plugged into existing language-image pretrained models, without undermining their original parameters and performance.
    \item We design a video-specific  prompting technique to yield instance-level textual representation automatically. It leverages video content information to enhance the textual prompt generation. 
    \item Our work presents a new way of expanding existing large-scale language-image pretrained models for general video recognition and other potential video tasks. Extensive experiments demonstrate the superiority and good generalization ability of our method under various learning configurations. 
\end{itemize}

\section{Related Work}

\textbf{Visual-language Pretraining.} Visual-language pretraining has achieved remarkable progress over the past few years \cite{sun2019videobert,sun2019learning,miech2019howto100m,zhu2020actbert}.
In particular, contrastive language-image pretraining demonstrates very impressive ``zero-shot" transfer and generalization capacities \cite{clip,align,florence}. One of the most representative works is the recent CLIP \cite{clip}. A large amount of follow-up works have been proposed to leverage the pretrained models for downstream tasks. For example, CoOp \cite{zhou2021coop}, CLIP-Adapter \cite{gao2021clipadapter} and Tip-Adapter \cite{zhang2021tip} use the pretrained CLIP for improving the few-shot transfer, while PointCLIP \cite{zhang2021pointclip} and DenseCLIP \cite{rao2021denseclip,zhou2021denseclip} transfer the knowledge to point cloud understanding and dense prediction, respectively. VideoCLIP \cite{xu2021videoclip} extends the image-level pretraining to video by substituting the image-text data with video-text pairs \cite{miech2019howto100m}. However, such video-text pretraining is computationally expensive and requires a large amount of curated video-text data which is not easy to acquire. In contrast, our method directly adapts the existing pretrained model to video recognition, largely saving the training cost.

There are two concurrent works mostly related to ours. One is ActionCLIP \cite{wang2021actionclip}, while the other is \cite{ju2021prompting}. Both of them introduce visual-language pretrained models to video understanding. ActionCLIP proposes a ``pretrain, prompt and finetune'' framework for action recognition, 
while \cite{ju2021prompting} proposes to optimize a few random vectors for adapting CLIP to various video understanding tasks. In contrast, our method is more general. It supports adapting various language-image models, such as CLIP and Florence \cite{florence}, from image to video. Moreover, we propose a lightweight and efficient cross-frame attention module for video temporal modeling, while presenting a new video-specific text prompting scheme.

\textbf{Video Recognition.} One key factor to build a robust video recognition model is to exploit the temporal information. Among many methods, 3D convolution is widely used~\cite{tran2015learning,tran2018closer,qiu2017P3D,xie2018rethinking}, while it suffers from high computational cost. For efficiency purposes, some studies~\cite{tran2018closer,qiu2017P3D,xie2018rethinking} factorize convolutions across spatial and temporal dimensions, while others insert the specific temporal modules into 2D 
CNNs~\cite{lin2019tsm,li2020tea,liu2021tam}. Nevertheless, the limited receptive field of CNNs gives the rise of transformer-based methods \cite{arnab2021vivit,timesformer2021,liu2021video,Fan_2021_ICCV,yan2022multiview}, which achieve very promising performance recently. However, these transformer-based methods are either computationally intensive or insufficient in exploiting the temporal information. For example, ViViT~\cite{arnab2021vivit} disregards the temporal information in the early stage. Video Swin~\cite{liu2021video} utilizes 3D attention while having high computational cost. 

The temporal modeling scheme in our method shares a similar spirit with the recent proposed video transformers, \emph{i.e.}, VTN \cite{neimark2021VTN}, ViViT \cite{arnab2021vivit}, and AVT \cite{avt}. They all use a frame-level encoder followed by a temporal encoder, but our method has two fundamental differences. 1) In  \cite{neimark2021VTN,arnab2021vivit,avt}, each frame is encoded separately, resulting in no temporal interaction before final aggregation. This late fusion strategy does not fully make use of the temporal cues. By contrast, our method replaces the spatial attention with the proposed cross-frame attention, which allows global spatio-temporal modeling for all frames. 2) Similar to previous works \cite{liu2021video,Fan_2021_ICCV,feichtenhofer2019slowfast,timesformer2021}, both ViViT \cite{arnab2021vivit} and VTN \cite{neimark2021VTN} adopt a dense temporal sampling strategy and ensemble the predictions of multiple views at inference, which is time-consuming. On the contrary, we empirically analyze different sampling methods for late fusion, and demonstrate that a sparse sampling is good enough, achieving better performance with fewer FLOPs than the dense strategy, as verified in Sec.~\ref{sec:ablation} (Analysis). 

\section{Approach}
In this section, we present our proposed framework in detail. First, we briefly overview our video-text framework in Sec.~\ref{sec:overview}. Then, we depict the architecture of the video encoder, especially for the proposed cross-frame attention in Sec.~\ref{sec:encoder}. Finally, we introduce a video-specific prompting scheme in Sec.~\ref{sec:text}.

\begin{figure}[tb!]
\centering
\includegraphics[scale=0.33]{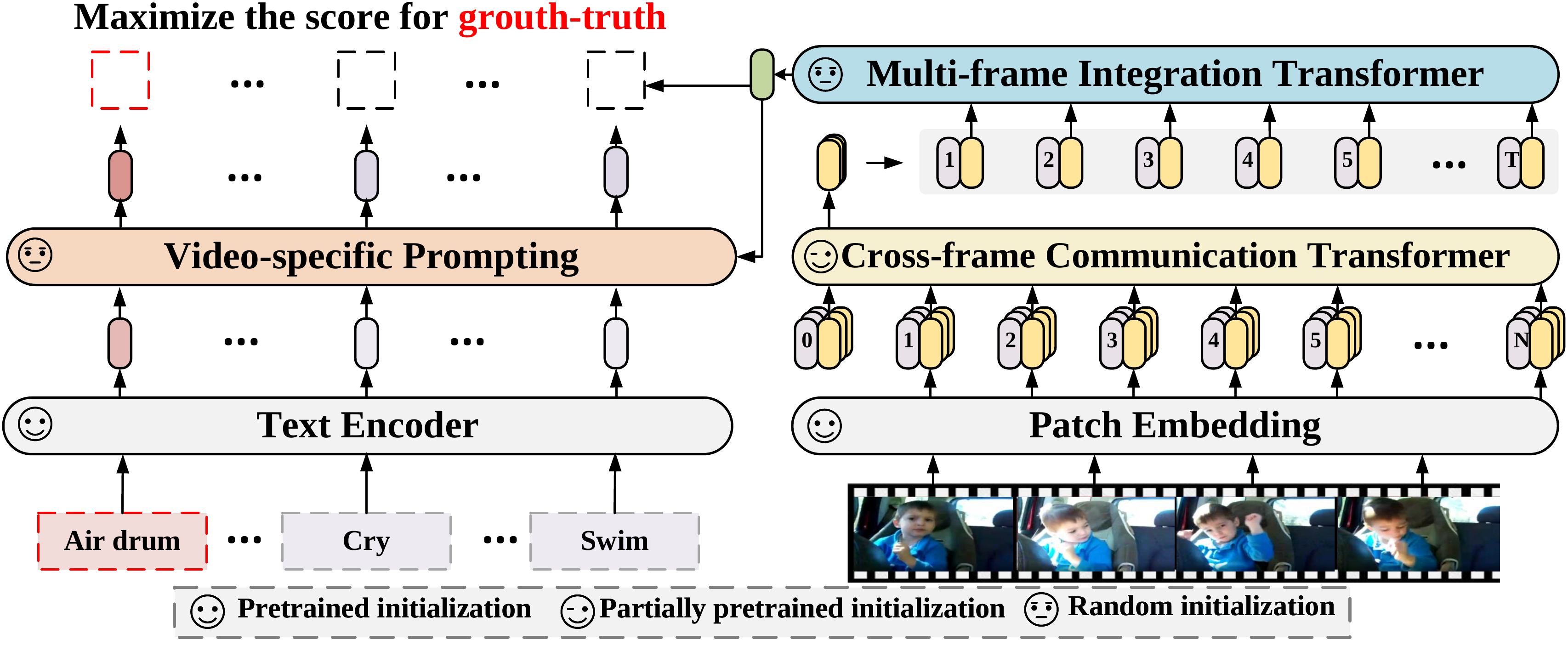}
\caption{An overview of our framework. The details are elaborated in Sec. \ref{sec:overview}.
}
\label{fig:overview}
\end{figure}

\subsection{Overview}\label{sec:overview}

Most prior works in video recognition learn discriminative feature embeddings supervised by a one-hot label \cite{arnab2021vivit,timesformer2021,feichtenhofer2019slowfast,wang2016tsn}. While in this work, inspired by the recent contrastive language-image pretraining \cite{clip,align,florence}, we propose to use text as the supervision, since the text provides more semantic information. As shown in Fig.~\ref{fig:overview}, our method learns to align the video representation and its corresponding text representation by jointly training a video encoder and a text encoder. Rather than pretraining a new video-text model from scratch, our method is built upon prior language-image models and expands them with video temporal modeling and video-adaptive textual prompts. Such a strategy allows us to fully take advantage of existing large-scale  pretrained models while transferring their powerful generalizability from image to video in a seamless fashion. 

Formally, given a video clip $V \in \mathcal{V}$ and a text description $C \in \mathcal{C}$, where $\mathcal{V}$ is a set of videos and $\mathcal{C}$ is a collection of category names, we feed the video $V$ into the video encoder $f_{\theta_v}$ and the text $C$ into the text encoder $f_{\theta_c}$ to obtain a video representation ${\bf{v}}$ and a text representation ${\bf{c}}$ respectively, where 
\begin{equation}
{\bf{v}} = f_{\theta_v}(V), \ \  {\bf{c}} = f_{\theta_c}(C).
\end{equation}
Then, a video-specific prompt generator $f_{\theta_p}$ is employed to yield instance-level textual representation for each video. It takes the video representation ${\bf{v}}$ and text representation ${\bf{c}}$ as inputs, formulated as 
\begin{equation}
{\hat{\bf{c}}} = f_{\theta_p}(\bf{c}, \bf{v}).\vspace{-.15cm}
\end{equation}
Finally, a cosine similarity function {$\mathrm{sim}$}({\bf{v}}, {$\hat{\bf{c}}$}) is utilized to compute the similarity between the visual and textual representations:
\begin{equation}
    {\rm sim}({\bf{v}},\hat{\bf{c}}) = \frac{\langle {\bf{v}}, \hat{\bf{c}}\rangle}{\left\|{\bf{v}}\right\| \left\|\hat{\bf{c}}\right\|}.\vspace{-.15cm}
\label{eq:cosine_similarity}
\end{equation}

\noindent The goal of our method is to maximize the ${\rm sim}({\bf{v}}, \hat{\bf{c}})$ if $V$ and $C$ are matched and otherwise minimize it. 

\begin{figure}[tb!]
\centering
\includegraphics[scale=0.38]{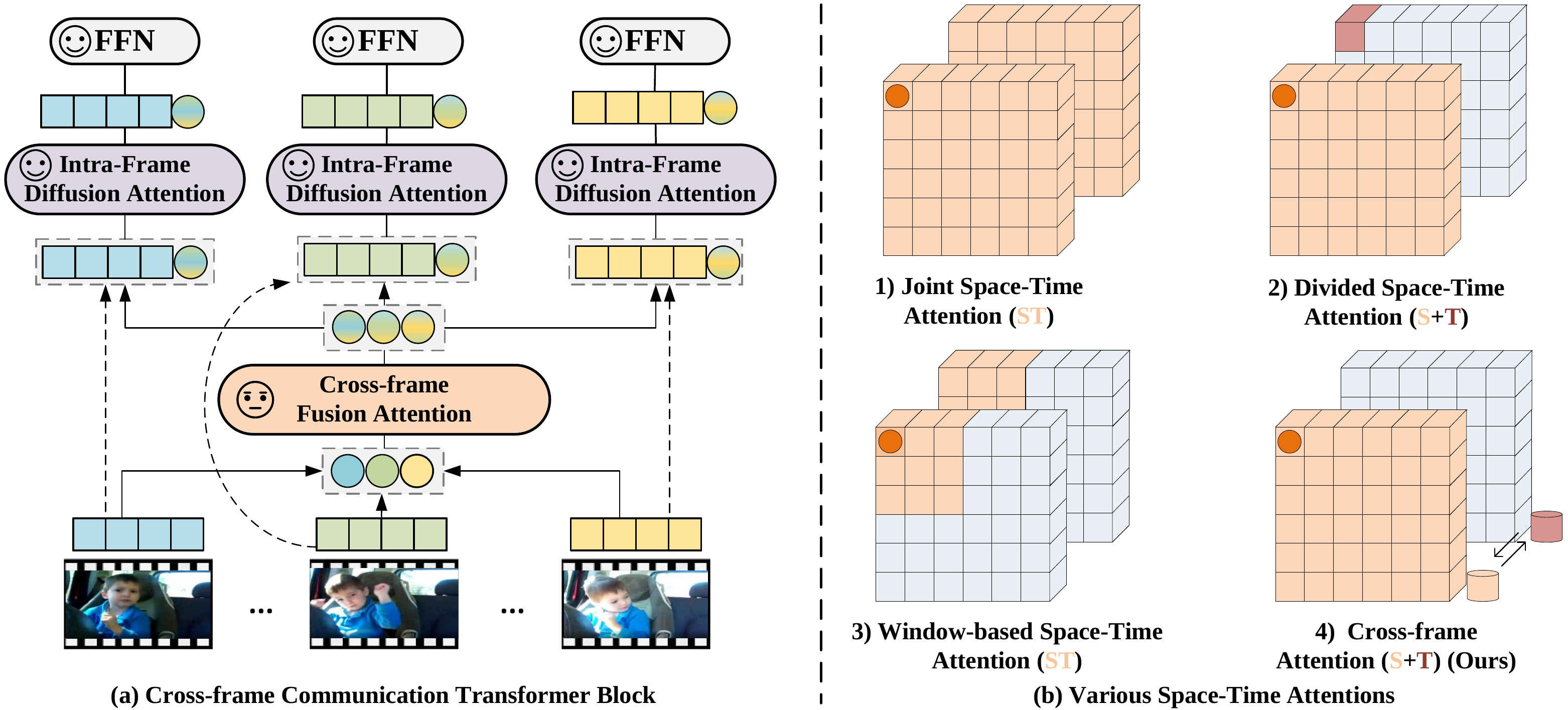}
\caption{ (a) Cross-frame Attention. (b) compares different space-time attention mechanisms used in existing video transformer backbones \cite{arnab2021vivit,timesformer2021,liu2021video}.}
\label{fig:attention}
\end{figure}
\subsection{Video Encoder}\label{sec:encoder}

Our proposed video encoder is composed of two cascaded vision transformers: a cross-frame communication transformer and a multi-frame integration transformer. The cross-frame transformer takes raw frames as input and provides a frame-level representation using a pretrained language-image model, while allowing information exchange between frames. The multi-frame integration transformer then simply integrates the frame-level representations and outputs video features.

Specifically, given a video clip $V \in \mathbb{R}^{T\times H \times W \times 3}$
of $T$ sampled frames with $H$ and $W$ denote the spatial resolution, following ViT \cite{dosovitskiy2020image}, the $t$-th frame is divided into $N$ non-overlapping patches $\{{\bf{x}}_{t,i}\}_{i=1}^{N} \in \mathbb{R}^{P^2\times3}$ 
with each of size $P \times P$ pixels, where $t\in\{1,\cdots,T\}$ denotes the temporal index, and $N$=$HW/P^2$. The patches $\{{\bf{x}}_{t,i}\}_{i=1}^{N}$ are then embedded into patch embeddings using a linear projection ${\bf{E}} \in \mathbb{R}^{3P^2 \times D}$. After that, we prepend a learnable embedding ${\bf{x}}_{class}$ to the sequence of embedded patches, called [\texttt{class}] token. Its state at the output of the encoder serves as the frame representation. The input of the cross-frame communication transformer at the frame $t$ is denoted as:

\begin{equation}
{\bf{z}}^{(0)}_{t} = [{\bf{x}}_{class}, {\bf{Ex}}_{t,1}, {\bf{Ex}}_{t,2}, \cdots,{\bf{Ex}}_{t,N}] + {\bf{e}}^{spa}, 
\end{equation}
where ${\bf{e}}^{spa}$ represents the spatial position encoding.

Then we feed the patch embeddings into an $L_c$-layer Cross-frame Communication Transformer (CCT) to obtain the frame-level representation ${\bf{h}}_t$:
\begin{equation}
\begin{aligned}
    {\bf{z}}^{(l)}_{t} &= \operatorname{CCT}^{(l)}({{\bf{z}}^{(l-1)}_{t}}), ~ l = 1, \cdots, L_c\\
    {\bf{h}}_t &= {\bf{z}}^{(L_c)}_{t, 0},
\end{aligned}
\end{equation}

where $l$ denotes the block index in CCT, ${\bf{z}}^{(L_c)}_{t,0}$ represents the final output of the [\texttt{class}] token. CCT is built-up with the proposed cross-frame attention, as will be elaborated later.

At last, the $L_m$-layer Multi-frame Integration Transformer (MIT) takes all frame representation ${\bf{H}} = [{\bf{h}}_1, {\bf{h}}_2, \cdots, {\bf{h}}_T] $ as input and outputs the video-level representation ${\bf{v}}$ as following:
\begin{equation}
    {\bf{v}} = \operatorname{AvgPool}(\operatorname{MIT}({\bf{H}}+{\bf{e}}^{temp})),
\end{equation}
where $\operatorname{AvgPool}$ and ${\bf{e}}^{temp}$ denote the average pooling and temporal position encoding, respectively. We use standard learnable absolute position embeddings \cite{vaswani2017attention} for ${\bf{e}}^{spa}$ and ${\bf{e}}^{temp}$. The multi-frame integration transformer is constructed by the standard \textcolor{black}{multi-head self-attention} and feed-forward networks \cite{vaswani2017attention}.

\vspace{-0.4cm}
\subsubsection{Cross-frame Attention.} 
To enable a cross-frame information exchange, we propose a new attention module. It consists of two types of attentions, \emph{i.e.}, cross-frame fusion attention (CFA) and intra-frame diffusion attention (IFA), with a feed-forward network (FFN). We introduce a \textit{message token} mechanism for each frame to abstract, send and receive information, thus enabling visual information to exchange across frames, as shown in Fig.~\ref{fig:attention}(a). In detail, the message token ${\bf{m}}_t^{(l)}$ for the $t$-th frame at the $l$-th layer is obtained by employing 
a linear transformation on the 
[\texttt{class}] token ${\bf{z}}_{t,0}^{(l-1)}$.
This allows message tokens to abstract the visual information of the current frame.

Then, the cross-frame fusion attention (CFA) involves all message tokens to learn the global spatio-temporal dependencies of the input video.  
Mathematically, this process at $l$-th block can be expressed as:
\begin{equation}
    \hat{{\bf{M}}}^{(l)} = {\bf{M}}^{(l)} + \operatorname{CFA}({\operatorname{LN}(\bf{M}}^{(l)})),
    \label{equ:8}
\end{equation}
where ${\hat{\bf{M}}}^{(l)}=[\hat{\bf{m}}^{(l)}_1, \hat{\bf{m}}^{(l)}_2, \cdots, \hat{\bf{m}}^{(l)}_T]$ and $\operatorname{LN}$ indicates layer normalization \cite{layernorm}.

Next, the intra-frame diffusion (IFA) takes the frame tokens with the associated message token to learn visual representation, while the involved message token could also diffuse global spatio-temporal dependencies for learning. Mathematically, this process at $l$-th block can be formulated as:
\begin{equation}
    [\hat{{\bf{z}}}^{(l)}_t, \bar{\bf{m}}_t^{(l)}] = [{\bf{z}}^{(l-1)}_t, \hat{{\bf{m}}}_t^{(l)}] + \operatorname{IFA}(\operatorname{LN}([{\bf{z}}^{(l-1)}_t, \hat{{\bf{m}}}_t^{(l)}])),\vspace{-.1cm}
\end{equation}
where $[\cdot,\cdot]$ concatenates the features of frame tokens and message tokens.

Finally, the feed-forward network(FFN) performs on the frame tokens as:
\begin{equation}
    {\bf{z}}^{(l)}_t = \hat{{\bf{z}}}^{(l)}_t + \operatorname{FFN}(\operatorname{LN}(\hat{{\bf{z}}}^{(l)}_t)).\label{equ:10}
\end{equation}
Note that the message token is dropped before the $\operatorname{FFN}$ layer and does not pass through the next block, since it is generated online and used for frames communication within each block. Alternating the fusion and diffusion attentions through $L_c$ blocks, the cross-frame communication transformer (CCT) can encode the global spatial and temporal information of video frames. Compared to other space-time attention mechanisms \cite{arnab2021vivit,timesformer2021,liu2021video}, as presented in Fig.~\ref{fig:attention}(b), our proposed cross-frame attention models the global spatio-temporal information while greatly reducing the computational cost.

\textit{Initialization.} When adapting the pretrained image encoder to the video encoder, there are two key modifications. 1) The intra-frame diffusion attention (IFA) inherits the weights directly from the pretrained models, while the cross-frame fusion attention (CFA) is randomly initialized. 2) The multi-frame integration transformer is appended to the pretrained models with random initialization. 

\subsection{Text Encoder}\label{sec:text}
We employ the pretrained text encoder and expand it with a video-specific prompting scheme. The key idea is to use video content to enhance the text representation. Given a description $C$ about a video, the text representation $\bf{c}$ is obtained by the text encoder, where ${\bf{c}}=f_{\theta_c}(C)$. For video recognition, how to generate a good text description $C$ for each video is a challenging problem. Previous work, such as CLIP~\cite{clip}, usually defines textual prompts manually, such as ``\texttt{A photo of a \{label\}}". However, in this work, we empirically show that such manually-designed prompts do not improve the performance for video recognition (as presented in Tab. \ref{tab:prompt}). In contrast, we just use the ``\texttt{\{label\}}" as the text description $C$ and then propose a learnable text prompting scheme. 

\subsubsection{Video-specific prompting.} When understanding an image or a video, human can instinctively seek helps from  discriminative visual cues. For example, the extra video semantic information of ``\texttt{in the water}" will make it easier to distinguish ``\texttt{swimming}" from ``\texttt{running}". However, it is difficult to acquire such visual semantics in video recognition tasks, because 1) the datasets only provide the category names, such as ``swimming" and ``running", which are pre-defined and fixed; and 2) the videos in the same class share the identical category name, but their visual context and content are different. To address these issues, we propose a learnable prompting scheme to generate textual representation automatically. Concretely, we design a video-specific prompting module, which takes the video content representation $\bar{\bf{z}}$ and text representation $\bf{c}$ as inputs. Each block in the video-specific prompting module is consisting of a multi-head self-attention (MHSA) \cite{vaswani2017attention} followed by a feed-forward network to learn the prompts,
\begin{equation}
\begin{aligned}
    \bar{\bf{c}} &= {\bf{c}} + \operatorname{\textcolor{black}{MHSA}}({\bf{c}}, \bar{\bf{z}}), \\
    \tilde{\bf{c}} &= \bar{\bf{c}} + \operatorname{FFN}(\bar{\bf{c}}),
\end{aligned}
\end{equation}
where $\bf{c}$ is the text embedding, ${\bar{\bf{z}}} \in \mathbb{R}^{N\times d}$ is the average of $\{{\bf{z}}_{t}^{(L_c)}\}_{t=1}^{T}$ along the temporal dimension, and $\tilde{{\bf{c}}}$ is the video-specific prompts. We use text representation ${\bf{c}}$ as query and the video content representation $\bar{\bf{z}}$ as key and value. This implementation allow the text representation to extract the related visual context from videos. We then enhance the text embedding $\bf{c}$ with the video-specific prompts $\tilde{\bf{c}}$ as follows, $\hat{\bf{c}} = \bf{c} + \boldsymbol \alpha \tilde{\bf{c}}$, where $\alpha$ is a learnable parameter with an initial value of 0.1. The $\hat{\bf{c}}$ is finally used for classification in Eq.~(\ref{eq:cosine_similarity}).

\section{Experiments}
In this section, we conduct experiments on different settings, \emph{i.e.},  fully-supervised, zero-shot and few-shot, followed by the ablation studies of the proposed method.
\begin{table*}[tb!]

\centering
\caption{Comparison with state-of-the-art on Kinetics-400. 
We report the FLOPs and throughput per view. 
Throughput is measured using the GitHub repository of~\cite{liu2021swin} on a V100 GPU. 
$*$ indicates pretraining with a video-text collection. 
}
\resizebox{0.95\textwidth}{!}{

\begin{tabular}{lccccccc}
  \Xhline{1.0pt}
  Method & Pretrain & Frames & Top-1 & Top-5 & Views & FLOPs(G) & Throughput  \\
  \hline
  \multicolumn{7}{l}{\textit{Methods with random initialization}} \\
  MViTv1-B, 64×3~\cite{Fan_2021_ICCV} & - & 64 & 81.2 & 95.1 & 3 × 3 & 455 & 7 \\
  \hline
  \multicolumn{7}{l}{\textit{Methods with ImageNet pretraining}} \\
  Uniformer-B~\cite{li2022uniformer} & IN-1k & 32 & 83.0 & 95.4 & 4 × 3 & 259 & - \\ 
  TimeSformer-L~\cite{timesformer2021} & IN-21k & 96 & 80.7 & 94.7 & 1 × 3 & 2380 & 3 \\
  Mformer-HR~\cite{patrick2021keeping} & IN-21k & 16 & 81.1 & 95.2 & 10 × 3 & 959 & - \\
  Swin-L~\cite{liu2021video} & IN-21k & 32 & 83.1 & 95.9 & 4 × 3 & 604 & 6 \\
  Swin-L (384$\uparrow$)~\cite{liu2021video} & IN-21k & 32 & 84.9 & 96.7 & 10 × 5 & 2107 & - \\
  MViTv2-L (312$\uparrow$)~\cite{li2021mvitv2} & IN-21k  & 40 & 86.1 & 97.0 & 5 × 3 & 2828 & - \\
  \hline
  \multicolumn{7}{l}{\textit{Methods with web-scale image pretraining}} \\
  ViViT-H/16x2~\cite{arnab2021vivit} & JFT-300M & 32 & 84.8 & 95.8 & 4 × 3 & 8316 & - \\
  TokenLearner-L/10~\cite{ryoo2021tokenlearner} & JFT-300M & - & 85.4 & 96.3 & 4 × 3 & 4076 & - \\
  CoVeR~\cite{zhang2021co} & JFT-3B & - & \textcolor{blue}{\textbf{87.2}} & - & 1 × 3 & - & - \\
  \hline
  \multicolumn{7}{l}{\textit{Methods with web-scale language-image pretraining}} \\
  ActionCLIP-B/16~\cite{wang2021actionclip}  & CLIP-400M & 32 & 83.8 & 96.2 & 10 × 3 & 563 & - \\
  A6~\cite{ju2021prompting} & CLIP-400M  & 16 & 76.9 & 93.5 & - & - & - \\

  MTV-H~\cite{yan2022multiview} & WTS$^*$ & 32 & \textcolor{red}{\textbf{89.1}} & 98.2 & 4 × 3 & 3705 & - \\
  \hline
  X-Florence (384$\uparrow$)& FLD-900M & 8 & 86.2 & 96.6 & 4 × 3 & 2114 & 6 \\
  X-Florence & FLD-900M & 32 & 86.5 & 96.9 & 4 × 3 & 2822 & 2 \\
  X-CLIP-B/16 & IN-21k & 8 & 81.1 & 94.7 & 4 × 3 & 145 & 33 \\
  X-CLIP-B/32 & \multirow{6}{*}{\rotatebox{270}{CLIP-400M}} & 8 & 80.4 & 95.0 & 4 × 3 & 39 & 136 \\
  X-CLIP-B/32 & & 16 & 81.1 & 95.5 & 4 × 3 & 75 & 69 \\

  X-CLIP-B/16 &  & 8 & 83.8 & 96.7 & 4 × 3 & 145 &  33\\
  X-CLIP-B/16 & & 16 & 84.7 & 96.8 & 4 × 3 & 287 & 17 \\
  X-CLIP-L/14 &  & 8 & 87.1 & 97.6 & 4 × 3 & 658 & 8 \\
  X-CLIP-L/14 (336$\uparrow$) &  & 16 & \textcolor{green}{\textbf{87.7}} & 97.4 & 4 × 3 & 3086 & 2 \\

\Xhline{1.0pt}
\end{tabular}
\label{tab:k400}
}
\vspace{-.6cm}
\end{table*}

\subsection{Experimental Setup}\label{sec:datasets}

\textbf{Architectures and Datasets.} We expand CLIP and Florence to derive four variants: X-CLIP-B/32, X-CLIP-B/16, X-CLIP-L/14 and X-Florence, respectively. In detail, there are three parts in our framework: a cross-frame communication transformer followed by a multi-frame integration transformer and a text encoder. X-CLIP-B/32 adopts ViT-B/32 as parts of the cross-frame communication transformer, X-CLIP-B/16 uses ViT-B/16, while X-CLIP-L/14 employs ViT-L/14. For all X-CLIP variants, we use a simple 1-layer multi-frame integration transformer. For X-Florence, we stack a 4-layer multi-frame integration transformer. The number of the video-specific prompting blocks is set to 2 for all variants. We evaluate the efficacy of our method on four benchmarks: \textit{Kinetics-400$\&$600} \cite{k400,k600}, \textit{UCF-101} \cite{soomro2012ucf101} and \textit{HMDB-51} \cite{kuehne2011hmdb}. More details about architectures and datasets are provided in the \emph{supplementary materials}.

\begin{table}[tb!]
\caption{Comparison with state-of-the-art on Kinetics-600. }
\centering
\resizebox{0.95\textwidth}{!}{
 \begin{tabular}{lccccccc}
\Xhline{1.0pt}
  Method & Pretrain & Frames & Top-1 & Top-5 & Views & FLOPs & Throughput  \\
  \hline
  \multicolumn{7}{l}{\textit{Methods with random initialization}} \\
  MViT-B-24, 32×3~\cite{Fan_2021_ICCV} & - & 32 & 83.8 & 96.3 & 5 × 1 & 236 & - \\
  
  \hline
  \multicolumn{7}{l}{\textit{Methods with ImageNet pretraining}} \\
  Swin-L (384$\uparrow$)~\cite{liu2021video} & IN-21k & 32 & 86.1 & 97.3 & 10 × 5 & 2107 & -\\
  \hline
  \multicolumn{7}{l}{\textit{Methods with web-scale pretraining}} \\
  ViViT-L/16x2 320~\cite{arnab2021vivit} & JFT-300M & 32 & 83.0 & 95.7 & 4 × 3 & 3992 & -\\
  ViViT-H/16x2~\cite{arnab2021vivit}  & JFT-300M & 32 & 85.8 & 96.5 & 4 × 3 & 8316  & -\\
  TokenLearner-L/10~\cite{ryoo2021tokenlearner} & JFT-300M & - & 86.3 & 97.0 &  4 × 3 & 4076  & - \\
  Florence (384$\uparrow$)~\cite{florence} & FLD-900M & - & 87.8 & 97.8 & 4 × 3 & - & - \\
  CoVeR~\cite{zhang2021co} & JFT-3B & - & \textcolor{blue}{\textbf{87.9}} & - & 1 × 3 & - & - \\
  MTV-H~\cite{yan2022multiview} & WTS$^*$ & 32 & \textcolor{red}{\textbf{89.6}} & 98.3 & 4 × 3 & 3705 & - \\
  \hline
  X-CLIP-B/16 & \multirow{3}{*}{CLIP-400M} & 8 & 85.3 & 97.1 & 4 × 3 & 145 & 74  \\
  X-CLIP-B/16 &  & 16 & 85.8 & 97.3 & 4 × 3 & 287 & 40 \\

  X-CLIP-L/14 & & 8 & \textcolor{green}{\textbf{88.3}} & 97.7 & 4 × 3 & 658 & 20 \\

  \Xhline{1.0pt}
\end{tabular}
}
\vspace{-.4cm}
\label{tab:k600}
\end{table}
\subsection{Fully-supervised Experiments} \label{sec:full-supervised}
\textbf{Training and Inference.} We sample 8 or 16 frames with a sparse sampling method in fully-supervised experiments. All the expanded models are trained with 32 NVIDIA 32G V100 GPUs. The detailed hyperparameters are showed in the \emph{supplementary materials}.

\textbf{Results.} In Tab.~\ref{tab:k400}, we report the results on Kinetics-400 and compare our method with state-of-the-art under different pretraining, including random initialization, ImageNet-1k/21k \cite{deng2009imagenet} pretraining, web-scale image and language-image pretraining. We develop a family of models with different FLOPs by setting the number of sampled frames to $8$ or $16$. 

Compared to the methods pretrained on ImageNet-21k \cite{deng2009imagenet}, our X-CLIP-B/16$_{8f}$ (with 8 sampled frames) surpasses Swin-L~\cite{liu2021swin} by $+0.7\%$ with $4\times$ fewer FLOPs and running $5\times$ faster(as presented in Fig.~\ref{fig:throughputs_flops}). The underlying reason is that the 3D shift-window attention in Swin is inefficient. Also, our X-CLIP-L/14$_{8f}$ outperforms MViTv2-L \cite{li2021mvitv2} by $+1.0\%$ with $5\times$ fewer FLOPs. In addition, when pretraining the video encoder only on IN-21k, our method achieves higher performance than the recent TimeSformer-L \cite{timesformer2021} with fewer computation cost.

When compared to the methods using web-scale image pretraining, our X-CLIP is also competitive. For example, X-CLIP-L/14$_{8f}$ achieves $+2.3\%$ higher accuracy than ViViT-H \cite{arnab2021vivit} with $12\times$ fewer FLOPs. MTV-H \cite{yan2022multiview} achieves better results than ours, but it uses much more pretraining data. Specifically, MTV-H uses a 70M video-text dataset with about \textcolor{black}{17}B images, which are much larger than the 400M image-text data used in CLIP pretraining.

Moreover, compared to ActionCLIP \cite{wang2021actionclip}, which also adopts CLIP as the pretrained model, our X-CLIP-L/14$_{8f}$ is still superior, getting $+3.3\%$ higher accuracy with fewer FLOPs. There are two factors leading to the smaller FLOPs of our method. One is that X-CLIP does not use 3D attention like \cite{liu2021video} and has fewer layers. The other factor is that X-CLIP samples fewer frames for each video clip, such as 8 or 16 frames, \textcolor{black}{while ActionCLIP \cite{wang2021actionclip} using 32 frames.}

In addition, we report the results on Kinetics-600 in Tab.~\ref{tab:k600}. \textcolor{black}{Using only $8$ frames, our X-CLIP-B/16$_{8f}$ achieves a higher top-1 accuracy compared to ViViT-L, while using $27\times$ fewer FLOPs.}
More importantly, our X-CLIP-L/14$_{8f}$ achieves $88.3\%$ top-1 accuracy while using $5 \times$ fewer FLOPs compared to the current state-of-the-art method MTV-H \cite{yan2022multiview}. 

From the above fully-supervised experiments, we can observe that, our X-CLIP method achieves very competitive performance compared to prevailing video transformer models \cite{zhang2021co,florence,yan2022multiview,wang2021actionclip,ju2021prompting}. 
This mainly attributes to two factors.
1) The proposed cross-frame attention can effectively model temporal dependencies of video frames. 2) The joint language-image representation is successfully transferred to videos, unveiling its powerful generalization ability for recognition. 

\begin{table}[t!]
\begin{minipage}{0.48\columnwidth}
\caption{Zero-shot performances on HMDB51 \cite{kuehne2011hmdb} and UCF101 \cite{soomro2012ucf101}.}
\centering
\setlength{\tabcolsep}{1.0mm}{
\resizebox{0.85\textwidth}{!}{
\begin{tabular}{ccc} \toprule
	Method  & HMDB-51 & UCF-101 \\ \midrule
	MTE  \cite{xu2016multi}  &  19.7 $\pm$ 1.6 & 15.8 $\pm$ 1.3 \\
	ASR \cite{wang2017alternative} & 21.8 $\pm$ 0.9 & 24.4 $\pm$ 1.0 \\
	ZSECOC \cite{qin2017zero}  &  22.6 $\pm$ 1.2 & 15.1 $\pm$ 1.7 \\
	UR \cite{zhu2018towards} & 24.4 $\pm$ 1.6 & 17.5 $\pm$ 1.6 \\
	TS-GCN \cite{gao2019know}  & 23.2 $\pm$ 3.0 & 34.2 $\pm$ 3.1 \\
	E2E \cite{brattoli2020rethinking}  & 32.7 & 48 \\
	ER-ZSAR \cite{chen2021elaborative} & 35.3 $\pm$ 4.6 & 51.8 $\pm$ 2.9  \\
	ActionCLIP \cite{wang2021actionclip} & 40.8 $\pm$ 5.4 & 58.3 $\pm$ 3.4 \\\midrule
	\multirow{2}{*}{X-CLIP-B/16} & \textbf{44.6 $\pm$ 5.2} & \textbf{72.0 $\pm$ 2.3}\\
	& (\scriptsize\textcolor{ForestGreen}{\textbf{+3.8}}) & (\scriptsize\textcolor{ForestGreen}{\textbf{+13.7}})  \\
	\multirow{2}{*}{X-Florence} & \textbf{48.4 $\pm$ 4.9} & \textbf{73.2 $\pm$ 4.2}\\
	& (\scriptsize\textcolor{ForestGreen}{\textbf{+7.6}}) & (\scriptsize\textcolor{ForestGreen}{\textbf{+14.9}})  \\
\bottomrule
\end{tabular}
}
}
\label{tab:zeroshot1}
\end{minipage}\noindent\hfill
\begin{minipage}{0.48\columnwidth}  
\caption{Zero-shot performance on Kinetics-600 \cite{k600}.}
\centering
\setlength{\tabcolsep}{1mm}{
\resizebox{0.85\textwidth}{!}{
	\begin{tabular}{ccc} \toprule
		Method & Top-1 Acc. & Top-5 Acc. \\ \midrule
		DEVISE \cite{frome2013devise} & 23.8 $\pm$ 0.3 & 51.0 $\pm$ 0.6 \\
		ALE \cite{akata2015label}  & 23.4 $\pm$ 0.8 & 50.3 $\pm$ 1.4 \\
		SJE \cite{akata2015evaluation}  & 22.3 $\pm$ 0.6 & 48.2 $\pm$ 0.4 \\
		ESZSL \cite{romera2015embarrassingly} & 22.9 $\pm$ 1.2 & 48.3 $\pm$ 0.8 \\
		DEM \cite{zhang2017learning}  & 23.6 $\pm$ 0.7 & 49.5 $\pm$ 0.4 \\
		GCN \cite{ghosh2020all}  & 22.3 $\pm$ 0.6 & 49.7 $\pm$ 0.6 \\  
		ER-ZSAR \cite{chen2021elaborative} & 42.1 $\pm$ 1.4 & 73.1 $\pm$ 0.3 \\ \midrule
	    \multirow{2}{*}{X-CLIP-B/16} & \textbf{65.2 $\pm$ 0.4} & \textbf{86.1 $\pm$ 0.8} \\
	    & (\scriptsize\textcolor{ForestGreen}{\textbf{+23.1}}) &  (\scriptsize\textcolor{ForestGreen}{\textbf{+13.0}}) \\
	    \multirow{2}{*}{X-Florence} & \textbf{68.8 $\pm$ 0.9} & \textbf{88.4 $\pm$ 0.6} \\
	    & (\scriptsize\textcolor{ForestGreen}{\textbf{+26.7}}) &  (\scriptsize\textcolor{ForestGreen}{\textbf{+15.3}}) \\
		\bottomrule
	\end{tabular}
	}
}
\label{tab:zeroshot2}

\end{minipage}
\vspace{-.47cm}
\end{table}
\subsection{Zero-shot Experiments} \label{sec:zero-shot}

\textbf{Training and Inference.} We pretrain X-CLIP-B/16 on Kinetics-400 with $32$ frames. The single view inference is adopted. More details about the evaluation protocols are provided in the \emph{supplementary materials}. 

\textbf{Results.} Zero-shot video recognition is very challenging, because the categories in the test set are unseen to the model during training. We report the results in Tab.~\ref{tab:zeroshot1} and Tab.~\ref{tab:zeroshot2}. On HMDB-51~\cite{kuehne2011hmdb} and UCF-101~\cite{soomro2012ucf101} benchmarks, our X-CLIP outperforms the previous best results by $+3.8\%$ and $+13.7\%$ in terms of top-1 accuracy respectively, as reported in Tab.~\ref{tab:zeroshot1}. On Kinetics-600~\cite{k600} as presented in Tab.~\ref{tab:zeroshot2}, our X-CLIP outperforms the state-of-the-art ER-ZSAR~\cite{chen2021elaborative} by $+23.1\%$. Such remarkable improvements can be attributed to the proposed video-text learning framework, which leverages the large-scale visual-text pretraining and seamlessly integrates temporal cues and textual prompts. 

\subsection{Few-shot Experiments} \label{sec:few-shot}

\textbf{Training and Inference.} A general $K$-shot setting is considered, \emph{i.e.}, $K$ examples are sampled from each category randomly for training. We compare with some representative methods, \emph{i.e.}, TSM \cite{lin2019tsm}, TimeSformer \cite{timesformer2021} and Swin \cite{liu2021video}. More details about the comparison methods and evaluation protocols are provided in the \emph{supplementary materials}.

\textbf{Results.} Tab.~\ref{tab:fewshot} presents the results of K-shot learning. For the extreme case where $K$=2, we observe that for those single-modality methods, the performance drops significantly, demonstrating that over-fitting occurs due to the serious lack of data. In contrast, X-CLIP shows robustness by surpassing them with large margins. For example, X-CLIP-B/16 outperforms Swin-B by $+32.1\%$ and $+23.1\%$ in terms of top-1 accuracy on HMDB-51 and UCF-101 with K=2, respectively. Such large improvements are mainly due to the exploitation of the semantics in text representation. It further verifies the efficacy of transferring the knowledge of the pretrained language-image models to the few-shot models. We also observe that the performance gap between our method and others decreases as the sample size increases. It demonstrates increasing data can mitigate the over-fitting for other methods. Besides, it is noteworthy that the comparison of methods with CLIP pretraining and ImageNet pretraining is not fair enough. Hence, in Sec.~\ref{sec:ablation}, we provide an additional ablation analysis and verify the performance gains mainly comes from the use of textual information, rather than the CLIP pretraining. 
\begin{table*}[tb!]
\caption{Few-shot results. Top-1 accuracy is reported with $32$ frames.}
\centering
\resizebox{0.85\textwidth}{!}{

  \begin{tabular}{c|cccc|cccc}
  \Xhline{1.0pt}
  \multirow{2}{*}{Method}  &\multicolumn{4}{c|}{HMDB-51} & \multicolumn{4}{c}{UCF-101} \\
  \cline{2-9}
                         & $K$=$2$ & $K$=$4$ & $K$=$8$ & $K$=$16$ & $K$=$2$ & $K$=$4$ & $K$=$8$ & $K$=$16$  \\
  \hline
  TSM~\cite{lin2019tsm} & 17.5 & 20.9 & 18.4 & 31.0 & 25.3 & 47.0 & 64.4 & 61.0 \\
  TimeSformer~\cite{timesformer2021} & 19.6 & 40.6 & 49.4 & 55.4 & 48.5 & 75.6 & 83.7 & 89.4 \\
  
  Swin-B~\cite{liu2021video} & 20.9 & 41.3 & 47.9 & 56.1 & 53.3 & 74.1 & 85.8 & 88.7 \\
  \hline
  
  \multirow{2}{*}{X-CLIP-B/16} &  \textbf{53.0} & \textbf{57.3} & \textbf{62.8} & \textbf{64.0} & \textbf{76.4} & \textbf{83.4} & \textbf{88.3} & \textbf{91.4} \\
    
    & \scriptsize(\textcolor{ForestGreen}{\textbf{+32.1}}) & \scriptsize(\textcolor{ForestGreen}{\textbf{+16.0}}) & \scriptsize(\textcolor{ForestGreen}{\textbf{+13.4}}) & \scriptsize(\textcolor{ForestGreen}{\textbf{+7.9}}) & \scriptsize(\textcolor{ForestGreen}{\textbf{+23.1}}) & \scriptsize(\textcolor{ForestGreen}{\textbf{+7.8}}) & \scriptsize(\textcolor{ForestGreen}{\textbf{+2.5}}) & \scriptsize(\textcolor{ForestGreen}{\textbf{+2.0}})  \\
    \multirow{2}{*}{X-Florence} &  \textbf{51.6} & \textbf{57.8} & \textbf{64.1} & \textbf{64.2} & \textbf{84.0} & \textbf{88.5} & \textbf{92.5} & \textbf{94.8} \\
    
    & \scriptsize(\textcolor{ForestGreen}{\textbf{+30.7}}) & \scriptsize(\textcolor{ForestGreen}{\textbf{+16.5}}) & \scriptsize(\textcolor{ForestGreen}{\textbf{+14.7}}) & \scriptsize(\textcolor{ForestGreen}{\textbf{+8.1}}) & \scriptsize(\textcolor{ForestGreen}{\textbf{+30.7}}) & \scriptsize(\textcolor{ForestGreen}{\textbf{+12.9}}) & \scriptsize(\textcolor{ForestGreen}{\textbf{+6.7}}) & \scriptsize(\textcolor{ForestGreen}{\textbf{+5.4}})  \\

  \Xhline{1.0pt}
  \end{tabular}
\label{tab:fewshot}
}
\vspace{-.4cm}
\end{table*}

\subsection{Ablation and Analysis}\label{sec:ablation}
Unless stated otherwise, the fully-supervised experiments are performed on Kinectics-400, while the few-shot experiments are conducted on HMDB-51 with $K$=2. The zero-shot evaluation is on the first split of the validation set of UCF-101. We use X-CLIP-B/16$_{8f}$ with single-view inference in all experiments.

\noindent\textbf{Ablation.}\textit{The effects of the proposed components.} Tab.~\ref{tab:step} shows the performance evolution from the pretrained image CLIP to our expanded video X-CLIP. First, we design a simple baseline that averages the CLIP features of all video frames for classification, called CLIP-Mean. It uses the text supervision but does not utilize prompting technique. We can observe that equipping the original transformer in CLIP with our proposed  cross-frame communication mechanism, \emph{i.e.} Eq. (\ref{equ:8}-\ref{equ:10}), can improve the accuracy by $+1.2\%$. Then, appending 1-layer multi-frame integration transformer (MIT) can further improve the accuracy by $+0.5\%$. This illustrates that our X-CLIP framework can effectively leverage temporal cues in video clips. With the proposed video-specific prompting, X-CLIP can surpass the CLIP-Mean baseline by $+2.3\%$. It demonstrates that the video-specific prompting scheme can generate more discriminative textual representation. Meanwhile, additionally using multi-view inference can boost the performance by $+1.5\%$. Overall, with our proposed methods and all the techniques mentioned above, X-CLIP can boost the top-1 accuracy of the CLIP-Mean baseline from $80.0$\% to $83.8$\%.

\begin{table}[tb!]
\begin{minipage}{0.475\columnwidth}
\caption{Component-wise analysis of our X-CLIP  and other techniques.}
\centering
  \begin{tabular}{lc} \toprule[0.5pt]
    Components & Top-1.(\%)\\ 
    \midrule[0.5pt] 
    Baseline(CLIP-Mean) & 80.0\\
    + Cross-frame Communication & 81.2\scriptsize(\textcolor{ForestGreen}{+1.2}) \\
    + Multi-frame Integration & 81.7\scriptsize(\textcolor{ForestGreen}{+1.7})  \\
    + Video-specific Prompt & 82.3\scriptsize(\textcolor{ForestGreen}{+2.3}) \\
    \midrule[0.5pt]
    Techniques & \\
    + 4$\times$3-views Inference & 83.8\scriptsize(\textcolor{ForestGreen}{+3.8})\\
    \bottomrule[0.5pt]
  \end{tabular}
  
\label{tab:step} 
\end{minipage}\noindent\hfill
\begin{minipage}{0.475\columnwidth}  
\centering
\caption{Ablation study on which part to finetune. \cmark means finetuning. The CUDA memory is calculated on 2 video inputs, each containing 8 frames.}
\centering
  \begin{tabular}{cc|ccc|c}
  \toprule[0.5pt]
  Visual & Text & Zero. & Few. & Fully. & Mem.(G)\\
  \midrule
  \cmark & \cmark & \textbf{72.9} & \textbf{54.6} & \textbf{82.4} & 22 \\
  \cmark & \xmark & \textbf{70.0} & 50.8 & \textbf{82.3} & 6 \\
  \xmark & \cmark & 66.8 & \textbf{53.4} & 79.3 & 20 \\
  \xmark & \xmark & 64.2 & 47.3 & 79.1 & 4 \\
  \bottomrule[0.5pt]
  \end{tabular}
\label{tab:frozen}
\end{minipage}
\vspace{-.6cm}
\end{table} 
\begin{table}[tb]
\begin{minipage}{0.475\columnwidth}
\centering
\caption{Ablation study on the effect of the text information. }
\centering

  \begin{tabular}{c|ccc}
  \toprule[0.5pt]
  Method & Zero-shot & Few-shot & Fully.\\
  \midrule
  w/o text & / & 32.0 & 81.6   \\
  w/ text & \textbf{70.0} & \textbf{50.8}\scriptsize(\textcolor{ForestGreen}{+18.8}) & \textbf{82.3}\scriptsize(\textcolor{ForestGreen}{+0.7})   \\
  \bottomrule[0.5pt]
  \end{tabular}
\label{tab:text}
\end{minipage}\noindent\hfill
\begin{minipage}{0.475\columnwidth}  
\centering
\caption{Comparison with different prompting methods.}
\scalebox{0.8}{
	\begin{tabular}{c|ccc}
    \toprule[0.5pt]
    Method & Fully. & Few. & Zero. \\
    \midrule
    w/o prompt & 81.7 & 49.6 & 63.2 \\
    Ensemble. \cite{clip} & 81.7 & 49.6 & 63.9 \\
    Vectors. \cite{zhou2021coop}  & 82.0 & 49.9  & 63.2 \\
    \midrule
    Ours & \textbf{82.3}\scriptsize(\textcolor{ForestGreen}{+0.3}) &
    \textbf{50.8}\scriptsize(\textcolor{ForestGreen}{+0.9}) &
    \textbf{70.0}\scriptsize(\textcolor{ForestGreen}{+6.1}) \\
    \bottomrule[0.5pt]
    \end{tabular}
}
\label{tab:prompt}
\end{minipage}
\vspace{-0.6cm}
\end{table}

\textit{Which branch to finetune?} In order to demonstrate which branch should be finetuned when transferred to different downstream tasks, we separately freeze the parameters of the pretrained image and text encoder. Note that the randomly initialized parameters are always finetuned. From Tab.~\ref{tab:frozen}, we summarize the following observations. 1) For fully-supervised setting, finetuning the image encoder brings $+3.0\%$ improvements, while freezing the text encoder reduces the CUDA memory from 22G to 6G with minor performance loss. 2) For few-shot setting, we find the top-2 results are achieved by finetuning the text encoder. We conjecture the reason is that with few samples, the text encoder suffers less from the over-fitting than the over-parameterized image model. 3) For zero-shot setting, finetuning both the image and the text encoder achieves the best results.

\textit{The effects of text.} To evaluate the impact of text, we replace the text encoder with a randomly initialized fully-connected layer as the classification head. From Tab.~\ref{tab:text}, we can observe that, without the text branch, the model cannot adapt to zero-shot setting, because there is no data to initialize the head. For the few-shot and fully-supervised experiments, text information can bring $+18.8\%$ and $+0.7\%$ gains, respectively. This indicates the semantic information involved in text representation is beneficial to classification, especially for low-shot learning.

\textit{The effects of pretraining.} We further investigate the effects of pretraining when expanding the language-image models to video. We use ViT-B/16 pretrained on ImageNet-1k/21k as the video encoder in our framework. As represented in Tab.~\ref{tab:pretrain}, though the pretrained image encoder and text encoder are not in a joint embedding space, the model with IN-21k and IN-1k pretraining still achieve $79.8\%$ and $75.9\%$ top-1 accuracy on Kinectics-400, yet much inferior to the original CLIP large-scale pretraining ($82.3\%$).

\begin{table}[tb]

\begin{minipage}{0.37\columnwidth}  
\caption{Ablation study on the different pretraining.}
\centering
\scalebox{0.9}{
  \begin{tabular}{c|c|c}
  \toprule[0.5pt]
  \multirow{2}{*}{Pretrain} & Top-1. & Top-5. \\
  & (\%) & (\%) \\
  \midrule
  ImageNet-1k & 75.9 & 90.2    \\
  ImageNet-21k  & 79.8 & 94.0  \\
  \bottomrule[0.5pt]
  \end{tabular}
}
\label{tab:pretrain} 
\end{minipage}\noindent\hfill
\begin{minipage}{0.58\columnwidth}

\caption{Comparison of two sampling methods. }
\scalebox{0.9}{
	\begin{tabular}{c|c|c|c}
    \Xhline{0.5pt}
     \multirow{2}{*}{\#F} & \multirow{2}{*}{\diagbox[width=40pt]{Train}{Test}} & \multicolumn{2}{c}{multi-view $\rightarrow$ single-view} \\
     \cline{3-4}
     & & Dense & Sparse \\
    \hline
    \multirow{2}{*}{8} & Dense & $81.9 \rightarrow 77.8$\scriptsize(-4.1) & $82.4 \rightarrow 81.1$\scriptsize(-1.3)  \\
      & Sparse & $ 82.2 \rightarrow 77.3$\scriptsize(-4.9) & $\textbf{83.4} \rightarrow \textbf{82.3}$\scriptsize(\textcolor{ForestGreen}{-1.1}) \\
    \hline
    \multirow{2}{*}{32} & Dense & $ 82.8 \rightarrow 78.8$\scriptsize(-4.0) & $83.2 \rightarrow 83.0$\scriptsize(-0.2) \\
     & Sparse & $83.0 \rightarrow 77.9$\scriptsize(-5.1) & $\textbf{84.4} \rightarrow \textbf{84.2}$\scriptsize(\textcolor{ForestGreen}{-0.2})  \\
  \Xhline{0.5pt}
  \end{tabular}
}
\label{tab:sampling}
\end{minipage}
\vspace{-0.5cm}
\end{table} 

\noindent\textbf{Analysis.}\textit{Comparison with other prompting methods.} We compare with two existing methods in Tab.~\ref{tab:prompt}: prompt ensembling \cite{clip} with 16 handcraft templates and learnable vectors \cite{zhou2021coop} with length 16. It can be seen that our video-specific prompts outperforms others, especially in zero-shot setting ($+6.1\%$). This demonstrates the efficacy of our method, which generates more adaptive prompts and better textual representation for unseen videos. 

\textit{Dense v.s. sparse sampling.} We further explore what is the best sampling strategy for our method in Tab.~\ref{tab:sampling}. We find that the dense sampling does not perform well as in previous works \cite{liu2021video,feichtenhofer2019slowfast,arnab2021vivit}. In contrast, the sparse sampling best matches our method. Regardless of the number of frames and views, using sparse sampling both in training and inference achieves the best performance.

\textit{Single-view v.s. multi-view inference.} Although it can improve performance, multi-view inference takes relatively high computational cost, because the cost grows linearly with the number of views. In Tab.~\ref{tab:sampling}, we show that our multi-modality models with sparse sampling is robust to the number of views, \emph{i.e.}, single-view can achieve comparable performance to 10 temporal views. The underlying reason is the language-image models provide robust representation.

\section{Conclusion}
In this work, we present a simple approach that adapts the pretrained language-image models to video recognition. To capture the temporal information, we propose a cross-frame attention mechanism that explicitly exchanges information across frames. A video-specific prompting technique is designed to yield instance-level discriminative textual representation. Extensive experiments under three different learning scenarios demonstrate the effectiveness of our method. In future work, we
plan to extend our method to different video tasks beyond classification.

\clearpage
%
%
\bibliographystyle{splncs04}
\bibliography{egbib}

\title{Expanding Language-Image Pretrained Models for General Video Recognition \\ --------Supplementary Material--------}


\titlerunning{X-CLIP}
%

%
\author{ }
\authorrunning{B. Ni et al.}
\institute{ }
%

\maketitle
This supplementary material contains additional details of the main manuscript, and provides more experiment analysis. 
In Sec.~\ref{sec:architecture_detail}, we present the details of our proposed architectures and the comparison methods. Next, we elaborate the hyperparameters in Sec.~\ref{sec:hyper}. Then, we overview the four datasets and provide the evaluation protocols of our experiments in Sec.~\ref{sec:dataset}. Finally, we provide more experiment analysis in Sec.~\ref{sec:additional_exp}.

\section{Architecture Details}\label{sec:architecture_detail}
In this section, we elaborate the details of the proposed architectures in Sec.~\ref{sec:our_architecture} and the compared architectures in the few-shot experiments in Sec.~\ref{sec:comparison_method}.
\subsection{The proposed architectures}\label{sec:our_architecture}
X-CLIP-B/32 adopts ViT-B/32  ($L_c$=$12$, $N_h$=$12$, $d$=$768$, $p$=$32$) as parts of the cross-frame communication transformer, X-CLIP-B/16 uses ViT-B/16 ($L_c$=$12$, $N_h$=$12$, $d$=$768$, $p$=$16$), while X-CLIP-L/14 employs ViT-L/14 ($L_c$=$24$, $N_h$=$16$, $d$=$1,024$, $p$=$14$), where $L_c$ denotes the layers, $N_h$ refers to the number of attention heads, $d$ represents the embedding dimension and $p$ is the patch size. We use a simple 1-layer multi-frame integration transformer for all three X-CLIP variants ($L_m$=$1$, $N_h$=$8$ for X-CLIP-B while $N_h$=$12$ for X-CLIP-L). The text encoder is the same as in CLIP~\cite{clip}. For Florence, we replace the cross-frame communication transformer with the pretrained CoSwin-H \cite{florence} visual encoder. We stack a 4-layer multi-frame integration transformer on top of CoSwin-H. The text encoder is the same as in Florence \cite{florence}.
\subsection{Other compared architectures}\label{sec:comparison_method}
In few-shot experiments, we implemented the Video Swin~\cite{liu2021video}, TSM~\cite{lin2019tsm} and TimeSformer~\cite{timesformer2021} using MMAction2~\cite{2020mmaction2} library with the default hyperparameters. The TSM-R50 is initialized with ImageNet-1k pretraining, while the Video Swin-B and TimeSformer are initialized with ImageNet-21k pretraining. 

\begin{table}[tb!]
\center
\caption{The training hyperparameters for all the experiments.}
\setlength{\tabcolsep}{2mm}{
\begin{tabular}{lccc}
\toprule[0.5pt]
                              & Fully-sup. & Few-shot & Zero-shot \\ \midrule
\multicolumn{4}{l}{\textit{Optimisation}} \\
Optimizer & \multicolumn{3}{c}{AdamW} \\
Optimizer betas					& \multicolumn{3}{c}{(0.9, 0.98)} \\
Batch size 						& 256 & 64 & 256 \\
Learning rate schedule	& \multicolumn{3}{c}{cosine} \\
Linear warmup epochs	& \multicolumn{3}{c}{5} \\
Base learning rate			& 8e-6 & 2e-6 & 8e-6 \\
Minimal learning rate & 8e-8 & 2e-8 & 8e-8 \\
Epochs	& 30 & 50 & 10 \\
\midrule
\multicolumn{4}{l}{\textit{Data augmentation}} \\
RandomFlip & \multicolumn{3}{c}{0.5} \\
MultiScaleCrop	& \multicolumn{3}{c}{(1, 0.875, 0.75, 0.66)} \\
ColorJitter  & \multicolumn{3}{c}{0.8} \\
GrayScale & \multicolumn{3}{c}{0.2} \\
Label smoothing & \multicolumn{3}{c}{0.1} \\
Mixup & \multicolumn{3}{c}{0.8} \\
Cutmix & \multicolumn{3}{c}{1.0} \\
\midrule
\multicolumn{4}{l}{\textit{Other regularisation}} \\
Weight decay & \multicolumn{3}{c}{0.001} \\
\bottomrule[0.5pt]
\end{tabular}
}
\label{tab:training_hyperparameters}
\vspace{-0.5cm}
\end{table}
\section{Hyperparameter Details}\label{sec:hyper}
In this section, we present the elaborated training hyperparameters in Sec.~\ref{sec:train_hyper} and the hand-craft prompt templates in the Tab. 9 of the main manuscript in Sec.~\ref{sec:prompt}.
\subsection{Training Hyperparameters}\label{sec:train_hyper}

Tab.~\ref{tab:training_hyperparameters} presents the hyperparameters for our experiments, corresponding to Section 4.2-4.4 of the main manuscript. It is noteworthy that the learning rate of the randomly initialized parameters is $10\times$ higher than the base learning rate.
\vspace{-0.3cm}
\subsection{Hand-craft Prompt Templates}\label{sec:prompt}
In Tab. 9 of the main manuscript, we compare our video-specific prompting scheme with the existing prompt ensemble method \cite{clip} and demonstrate the superiority of our method. We construct 16 hand-craft templates~\cite{wang2021actionclip} totally. We randomly choose one template in each training iteration, and the result in inference is the average result of all templates. The complete list of templates is as follows: a photo of action \{label\}; a picture of action \{label\};
Human action of \{label\}; \{label\}, an action; \{label\}, this is an action; \{label\}, a video of action; Playing action of \{label\}; \{label\}; Playing a kind of action, \{label\}; Doing a kind of action, \{label\}; Look, the human is \{label\}; Can you recognize the action of \{label\}; Video classification of \{label\}; A video of \{label\}; The man is \{label\}; The woman is \{label\}.

\section{Datasets and Evaluation Protocols}\label{sec:dataset}
In this section, we overview the four datasets briefly in Sec.~\ref{sec:dataset_overview}. 
Then, we provide the evaluation protocols for different experiment settings, \emph{i.e.}, zero-shot, few-shot and fully-supervised in Sec.~\ref{sec:split_fully}-\ref{sec:split_zero}, respectively.

\subsection{Datasets Overview}\label{sec:dataset_overview}
\begin{itemize}
\item \textit{Kinetics-400$\&$600.} The Kinetics~\cite{k400,k600} dataset consists of 10-second video clips collected from YouTube. In particular, Kinetics-400~\cite{k400} consists of $\sim$240k training videos and $\sim$20k validation videos with $400$ classes, while Kinetics-600~\cite{k600} consists of $\sim$410k training videos and $\sim$29k validation videos from $600$ classes. 

\item \textit{UCF-101} \cite{soomro2012ucf101}. UCF-101 is a video recognition dataset for realistic actions, collected from YouTube, including $13,320$ video clips with $101$ action categories in total. There are three splits of the training and test data.

\item \textit{HMDB-51} \cite{kuehne2011hmdb}. It has around $7,000$ videos with $51$ classes, which is relatively small compared to UCF-101 and Kinetics. HMDB-51 has three splits of the training and test data.
\end{itemize}

\subsection{Fully-supervised Experiments}\label{sec:split_fully}
We conduct the fully-supervised experiments on Kinetics-400$\&$600. We use the complete training and validation sets for training and inference, respectively. During training, a sparse sampling strategy~\cite{wang2016tsn} is used. The number of frames is set to 8 or 16. We spatially scale the shorter side of each frame to 256 and take a 224 center crop.  Following~\cite{liu2021video,arnab2021vivit,timesformer2021}, we adopt the multi-view inference with 3 spatial crops and 4 temporal clips.

\subsection{Few-shot Experiments}\label{sec:split_few}
We randomly sample 2, 4, 8 and 16 videos from each class on UCF-101 and HMDB-51 for constructing the training set. For evaluation, we use the first split of the test set on UCF-101 and HMDB-51. We report the results with a single view of 32 frames.

\subsection{Zero-shot Experiments}\label{sec:split_zero}
We train X-CLIP-B/16 with 32 frames on Kinetics-400. The single-view inference is adopted for our method. The same as ~\cite{chen2021elaborative,clip}, we apply the following two evaluation protocols in our zero-shot experiments. 1) \textit{Evaluation for HMDB-51 and UCF-101}. Following~\cite{clip}, the prediction is conducted on the three splits of the test data, and we report the average top-1 accuracy and standard deviation. 2) \textit{Evaluation for Kinetics-600.} Following~\cite{chen2021elaborative}, the $220$ new categories outside Kinetics-400~\cite{k400} in Kinetics-600 are used for evaluation. The evaluation is conducted three times. For each iteration, we randomly sampled 160 categories for evaluation from the 220 categories in Kinetics-600.

\section{Additional Experiments Analysis}\label{sec:additional_exp}
In this section, we further compare different methods of adapting an image encoder to a video encoder in Sec.~\ref{sec:encoder}. Besides, we provide an analysis of aligning the ImageNet pretrained video encoder and the CLIP pretrained text encoder in Sec.~\ref{sec:fewshot}. Last, we further evaluate our proposed cross-frame communication transformer and multi-frame integration transformer on a simple single-modality classification setting in Sec.~\ref{sec:simple}. 
\begin{table}[tb!]
\vspace{-.2cm}
\centering
\caption{Comparison with different video encoders. The video encoders are adapted from ViT-B/16 \cite{clip}. The fully-supervised experiment is conducted on Kinetics-400 \cite{k400}. The few-shot(2-shot) experiment is conducted on HMDB-51, and zero-shot experiment is conducted on UCF-101 \cite{soomro2012ucf101}.}
\centering
  \begin{tabular}{c|ccc|c}
  \toprule
  Method & Zero-shot & Few-shot & Fully-supervised & FLOPs\\
  \midrule
  CLIP-One & 62.5 & 46.2 & 69.9 & 14\\
  CLIP-Joint & 69.3 & 41.3 & 82.1 & 184 \\
  \midrule
  X-CLIP & \textbf{70.0}\scriptsize(+0.7) & \textbf{50.8}\scriptsize(+4.6) & \textbf{82.3}\scriptsize(+0.2) & 145 \\
  \bottomrule
  \end{tabular}
\label{tab:joint_mean}
\end{table}
\subsection{Comparison with other video encoders adapted from images}\label{sec:encoder}
Researchers have proposed several ways of adapting an image encoder to a video encoder \cite{arnab2021vivit,timesformer2021}. We compare with two existing methods in Tab.~\ref{tab:joint_mean}.  The first method is named ``CLIP-One", in which we randomly sample one frame and feed it to the pretrained image encoder. The second method is named ``CLIP-Joint", where we apply the joint space-time attention \cite{arnab2021vivit} that simply forwards all spatio-temporal tokens extracted from the video through the image encoder. Although the CLIP-Joint also considers global spatio-temporal information in videos, it takes more computational overhead than our proposed X-CLIP. What is more, our method surpasses the CLIP-Joint by $+0.2\%$ and $+9.5\%$ in the fully-supervised and few-shot experiments, respectively. We conjecture the reasons are two-fold. 1) CLIP-Joint considers the joint spatio-temporal tokens and therefore breaks the customary input pattern of the pretrained image encoder, which may impede the representation ability. In contrast, our method maintains the input pattern of the pretrained image encoder via modeling frame-level information, thus leveraging the strong representation ability of the pretrained image encoder. 2) The joint space-time attention requires more training data and training time to converge than our method. 

\begin{table}[tb!]
\renewcommand{\arraystretch}{1.0}
    \centering
    \caption{Comparison between the multi-modal framework and single-modal framework under ImageNet pretraining.}
    \setlength{\tabcolsep}{1.5mm}{
    \begin{subtable}[t]{2in}
        \centering
        \begin{tabular}{c|cc}
          \toprule
          Method & Zero-shot & Few-shot \\
          \midrule
          w/o text & / & 39.4  \\
          w/ text & \textbf{62.8} & \textbf{50.7}\scriptsize(+11.3)  \\
          \bottomrule
          \end{tabular}
        \caption{ImageNet-21k pretraining.}\label{tab:21k_pretraining}
    \end{subtable}}
    \hfill
    \setlength{\tabcolsep}{1.5mm}{
    \begin{subtable}[t]{2in}
        \centering
        \begin{tabular}{c|cc}
          \toprule
          Method & Zero-shot & Few-shot \\
          \midrule
          w/o text & / & 10.8 \\
          w/ text & \textbf{58.0} & \textbf{46.0}\scriptsize(+35.2)  \\
          \bottomrule
          \end{tabular}
        \caption{ImageNet-1k pretraining.}\label{tab:1k_pretraining}
    \end{subtable}}
    \label{tab:alignment}
\vspace{-0.8cm}
\end{table}

\begin{table}[tb!]
\center
\caption{Evaluating the proposed architecture in the single-modality framework.}

\begin{tabular}{c|cc}
  \toprule
  Method & Top-1(\%) & Top-5(\%) \\
  \midrule
  ViT-B/32-Mean & 45.3 & 68.5 \\
  ViT-B/32 (Ours) & \textbf{47.8}\scriptsize(+2.5) & \textbf{71.8}\scriptsize(+3.3) \\
  \bottomrule
\end{tabular}

\label{tab:simple}

\end{table}    

\subsection{Can ImageNet pretrained video encoder align with CLIP pretrained text encoder?}\label{sec:fewshot}
We have demonstrated that the video encoder with ImageNet pretraining still achieves competitive performance on the fully-supervised experiment in Tab. 10 of the main manuscript. However, the two embedding spaces of the ImageNet pretrained visual encoder and CLIP pretrained text encoder are not well aligned, which raises a question: \textit{can we align the two embedding spaces without the web-scale joint pretraining, and then transfer the knowledge to zero-shot experiments?} To answer this question, we build an ImageNet pretrained video encoder and separate the text encoder from the pretrained CLIP. Then, we finetune the video encoder with the text supervision on Kinetics-400 to align the two embedding spaces. As a comparison, we also finetune a same video encoder but supervised by the discrete one-hot labels. Finally, we conduct the few-shot and zero-shot experiments using the two finetuned models to verify the transfer ability. The categories in few-shot and zero-shot experiments are not seen in finetuning. From Tab.~\ref{tab:alignment}, we can observe that the aligned model, \emph{i.e.,} the model supervised by text information, achieves superior performance and surpasses the unaligned model by a large margin. It indicates that the ImageNet pretrained video encoder can still align with the CLIP pretrained text encoder by an acquired finetuning process using limited samples. The results also show the generality and flexibility of our proposed framework.

\subsection{Evaluation of the proposed architectures in the single-modality framework}\label{sec:simple}
We further evaluate the proposed cross-frame communication transformer and multi-frame integration transformer on a simple classification setting, \emph{i.e.,} training from scratch with a single-modality framework on Kinetics-400. We use ViT-B/32$_{8f}$ as the backbone and adopt a fully-connected layer as the classification head. ViT-B/32-Mean averages the representation of all frames, while our method uses the cross-frame attention and stacks 1-layer multi-frame integration transformer on the top. We train both models 100 epochs with a learning rate $1\times10^{-4}$, and all the other hyperparameters are the same as in Tab.~\ref{tab:training_hyperparameters}. From Tab.~\ref{tab:simple}, it can be seen that our method outperforms the baseline $+2.5\%$ in terms of top-1 accuracy, which illustrates that our proposed architecture does not rely on pretraining and can help general video classification.

\end{document}